\documentclass{article}
\usepackage{arxiv}

\usepackage[utf8]{inputenc} 
\usepackage[T1]{fontenc}    
\usepackage{hyperref}       
\usepackage{url}            
\usepackage{booktabs}       
\usepackage{amsfonts}       
\usepackage{nicefrac}       
\usepackage{microtype}      
\usepackage{xcolor}         
\usepackage{algorithm}
\usepackage{algorithmic}
\usepackage{graphicx}
\usepackage{bm}
\usepackage{array,multirow}
\usepackage{float}
\usepackage{subfig}
\usepackage{comment}
\usepackage{amsmath}
\usepackage{pgfplots}
\usepackage{tikz}
\usetikzlibrary{calc,arrows}
\usepackage{color, colortbl}
\DeclareMathOperator{\Tr}{Tr}
\definecolor{LightCyan}{rgb}{0.88,1,1}

\title{Adaptive Dimension Reduction and Variational Inference for Transductive Few-Shot Classification}

\author{%
  Yuqing Hu \\
  Orange Labs, Cesson-S\'evign\'e, France \\
  IMT-Atlantique, Brest, France \\
  \texttt{yuqing.hu@imt-atlantique.fr} \\
   \And
   St\'ephane Pateux\\
   Orange Labs, Cesson-S\'evign\'e, France \\
   \texttt{stephane.pateux@orange.com} \\
   \AND
   Vincent Gripon \\
   IMT-Atlantique, Brest, France \\
   \texttt{vincent.gripon@imt-atlantique.fr} \\
}

\begin{document}

\maketitle

\begin{abstract}
  Transductive Few-Shot learning has gained increased attention nowadays considering the cost of data annotations along with the increased accuracy provided by unlabelled samples in the domain of few shot. Especially in Few-Shot Classification (FSC), recent works explore the feature distributions aiming at maximizing likelihoods or posteriors with respect to the unknown parameters. Following this vein, and considering the parallel between FSC and clustering, we seek for better taking into account the uncertainty in estimation due to lack of data, as well as better statistical properties of the clusters associated with each class. Therefore in this paper we propose a new clustering method based on Variational Bayesian inference, further improved by Adaptive Dimension Reduction based on Probabilistic Linear Discriminant Analysis. Our proposed method significantly improves accuracy in the realistic unbalanced transductive setting on various Few-Shot benchmarks when applied to features used in previous studies, with a gain of up to $6\%$ in accuracy. In addition, when applied to balanced setting, we obtain very competitive results without making use of the class-balance artefact which is disputable for practical use cases. We also provide the performance of our method on a high performing pretrained backbone, with the reported results further surpassing the current state-of-the-art accuracy, suggesting the genericity of the proposed method.                   
\end{abstract}

\section{Introduction}
\label{introduction}
Few-shot learning, and in particular Few-Shot Classification, has become a subject of paramount importance in the last years with a large number of methodologies and discussions. Where large datasets continuously benefit from improved machine learning architectures, the ability to transfer this performance to the low-data regime is still a challenge due to the high uncertainty posed using few labels. In more details, there are two main types of FSC tasks. In \emph{inductive} FSC~\cite{DBLP:conf/iclr/AntoniouES19, snell2017prototypical, ye2020few, rizve2021exploring}, the situation comes to its extremes with only a few data samples available for each class, leading sometimes to completely intractable settings, such as when facing a black dog on the one hand and a white cat on the other hand. In \emph{transductive} FSC, additional unlabelled samples are available for prediction, leading to improved reliability and more elaborate solutions~\cite{lee2021few, lazarou2021iterative, baik2021meta}.

\def\connect#1#2#3{%
\path let
  \p1 = ($(#2)-(#1)$),
  \n1 = {veclen(\p1)},
  \n2 = {atan2(\x1,\y1)} 
in
  (#1) -- (#2) node[#3, midway, sloped, shading angle=\n2+90, minimum width=\n1, inner sep=0pt, #3] {};
}

\begin{figure}[h]
    \centering
    \subfloat[Few-Shot task]{\includegraphics[width=0.225\linewidth]{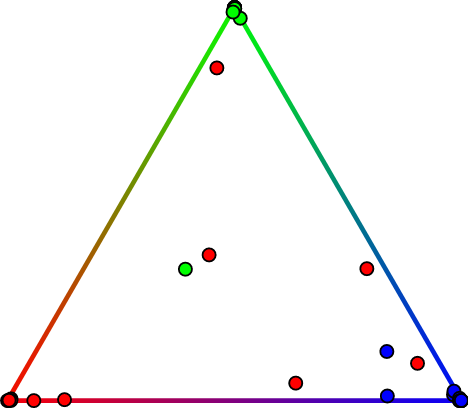}} \quad
    \subfloat[Initialization]{\includegraphics[width=0.225\linewidth]{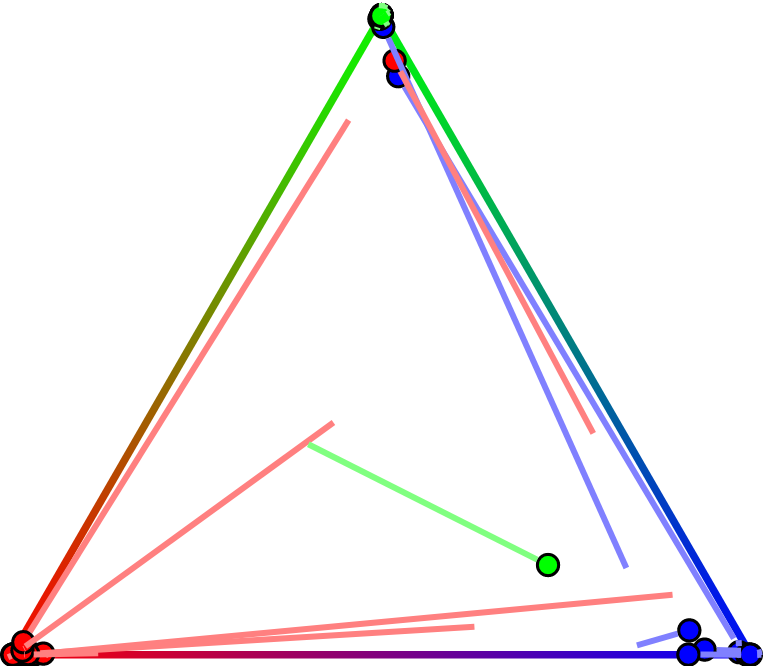}} \quad
    \subfloat[PLDA and VB inference]{\includegraphics[width=0.45\linewidth]{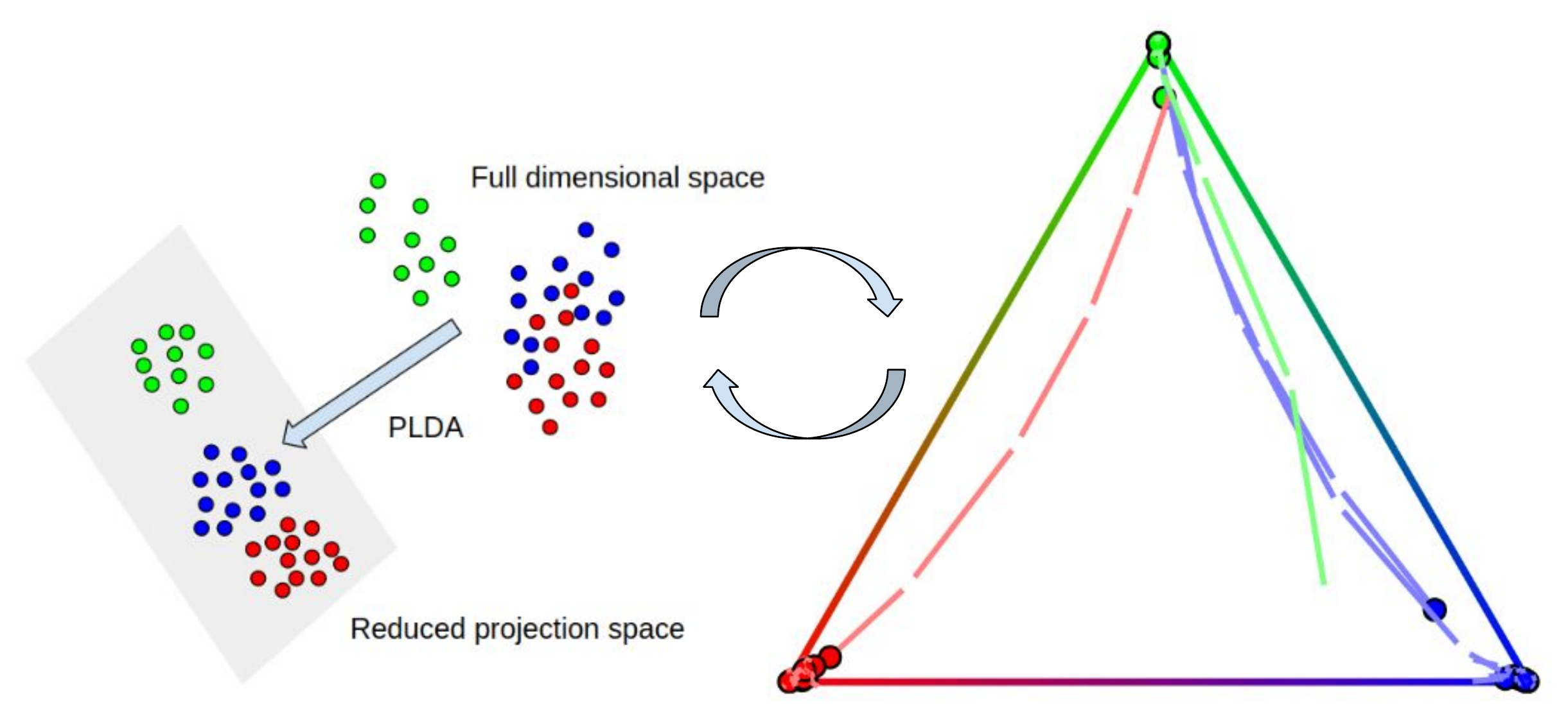}} 
    \caption{Summary of the proposed method. Here we illustrate a 3-way classification task in a standard 2-simplex using soft-classification probabilities. Trajectories show the evolution across iterations. For a given Few-Shot task which nearest-class-mean probabilities are depicted in (a), a Soft-KMEANS clustering method is performed in (b) to initialize $o_{nk}$ (see Alg.~\ref{alg:bavardage}). Then in (c) an iteratively refined Variational Bayesian (VB) model with Adaptive Dimension Reduction using Probabilistic Linear Discriminant Analysis (PLDA) is applied to obtain the final class predictions.}
    \label{fig:figure}
\end{figure}

Inductive FSC is likely to occur when data acquisition is difficult or expensive, or when categories of interest correspond to rare events. Transductive FSC is more likely encountered when data labeling is expensive, for fast prototyping of solutions, or when the categories of interest are rare and hard to detect. Since the latter correspond to situations where it is possible to exploit, at least partially, the distribution of unlabelled samples, the trend evolved to using potentially varying parts of this additional source of information. With most standardized benchmarks using very limited scope of variability in the generated Few-Shot tasks, this even came to the point the best performing methods are often relying on questionable information, such as equidistribution between the various classes among the unlabelled samples, that is unlikely realistic in applications.

This limitation of benchmarking for transductive FSC has recently been discussed in~\cite{veilleux2021realistic}. In this paper, the authors propose a new way of generating transductive FSC benchmarks where the distribution of samples among classes can drastically change from a Few-Shot generated task to the next one. Interestingly, they showed the impact of generating class imbalance on the performance on various popular methods, resulting in some cases in drops in average accuracy of more than 10\%.

A simple way to reach state-of-the-art performance in transductive FSC consists in extracting features from the available samples using a pretrained backbone deep learning architecture, and then using semi-supervised clustering routines to estimate samples distribution among classes. Due to the very limited number of available samples, distribution-agnostic clustering algorithms are often preferred, such as K-MEANS or its variants~\cite{moon1996expectation, lichtenstein2020tafssl, DBLP:conf/iclr/RenTRSSTLZ18} or mean-shift~\cite{comaniciu2002mean} for instance.

In this paper, we are interested in showing it is possible to combine data reduction with statistical inference through a Variational Bayesian (VB)~\cite{fox2012tutorial} approach. Here, data reduction helps considerably reduce the number of parameters to infer, while VB provides more flexibility than the usual K-Means methods. Interestingly, the proposed approach can easily cope with standard equidistributed Few-Shot tasks or the unbalanced ones proposed in~\cite{veilleux2021realistic}, defining a new state-of-the-art for five popular transductive Few-Shot vision classification benchmarks.

Our claims are the following:
\begin{itemize}
    \item We introduce a novel semi-supervised clustering algorithm based on VB inference and Probabilistic Linear Discriminant Analysis (PLDA),
    \item We demonstrate the general utility of our proposed method being able to improve accuracy in a variety of deep learning models and settings,
    \item We show the ability of the proposed method to reach state-of-the-art transductive FSC performance on multiple vision benchmarks (balanced and unbalanced).
\end{itemize}

\section{Related work}
\label{related}

There are two main frameworks in the field of FSC: 1) only one unlabelled sample is processed at a time for class predictions, which is called inductive FSC, and 2) the entire unlabelled samples are available for further estimations, which is called transductive FSC. Inductive methods focus on training a feature extractor that generalizes well the embedding in a feature sub-space, they include meta learning methods such as~\cite{finn2017model, liu2020prototype, baik2021meta, vinyals2016matching, oreshkin2018tadam, sung2018learning} that train a model in an episodic manner, and transfer learning methods~\cite{DBLP:conf/iclr/ChenLKWH19, mangla2020charting, ziko2020laplacian, NEURIPS2020_196f5641, bendou2022easy, rizve2021exploring} that train a model with a set of mini-batches. Recent state-of-the-art works on inductive FSC ~\cite{ye2020few, zhang2020deepemd, wertheimer2021few, kang2021relational} combine the above two strategies and propose a transfer based training used as model initialization, followed by an episodic training that adapts the model to better fit the Few-Shot tasks. 

Transductive methods are becoming more and more popular thanks to their better performance due to the use of unlabelled data, as well as their utility in situations where data annotation is costly. Early literature of this branch operates on a class-balanced setting where unlabelled instances are evenly distributed among targeted classes. Graph-based methods~\cite{gidaris2019generating, chen2021hierarchical, yang2020dpgn, kim2019edge} make use of the affinity among features and propose to group those that belong to the same class. More recent works such as~\cite{hu2021leveraging} propose methods based on Optimal Transport that realizes sample-class allocation with a minimum cost. While effective, these methods often require class-balanced priors to work well, which is not realistic due to the arbitrary unknown query set.  In~\cite{veilleux2021realistic} the authors put forward a novel unbalanced setting that composes a query set with unlabelled instances sampled following a Dirichlet distribution, injecting more imbalance for predictions.  

In this paper we propose a clustering method to solve transductive FSC, where the aim is to estimate cluster parameters giving high predictions for unlabelled samples. Under Gaussian assumptions, previous works~\cite{lichtenstein2020tafssl, DBLP:conf/iclr/RenTRSSTLZ18} have utilised algorithms such as Expectation Maximization~\cite{dempster1977maximum} (EM), with the goal of maximizing likelihoods or posteriors with respect to the parameters for a cluster, with the hidden variables marginalized. However, this may not be the most suitable way due to the scarcity of available data in a given Few-Shot task, which increases the level of uncertainty for cluster estimations. Therefore, in this paper we propose a Variational Bayesian (VB) approach, in which we regard some unknown parameters as hidden variables in order to inject more flexibility into the model, and we try to approximate the posterior of the hidden variables by a variational distribution.

As models with too few labelled samples often give too much randomness for a cluster to be stably reckoned, they often require the use of feature dimension reduction techniques to stabilize cluster estimations. Previous literature such as~\cite{lichtenstein2020tafssl} applies a PCA method that reduces dimension in a non-supervised manner, and~\cite{DBLP:conf/iclr/CaoLF20} proposes a modified LDA during backbone training that maximizes the ratio of inter/intra-class distance. In this paper we propose to use Probabilistic Linear Discriminant Analysis~\cite{ioffe2006probabilistic} (PLDA) that 1) is applied on extracted features, 2) fits data more desirably into distribution assumptions, and 3) is semi-supervised in combination of a VB model. We integrate PLDA into the VB model in order to refine the reduced space through iterations.

\section{Methodology}
\label{methodology}

In this section, we firstly present the standard setting in transductive FSC, including the latest unbalanced setting proposed by~\cite{veilleux2021realistic} where unlabelled samples are non-uniformly distributed among classes. Then we present our proposed method combining PLDA and VB inference.

\subsection{Problem formulation}
\label{formulation}

Following other works in the domain, our proposed method is operated on a feature space obtained from a pre-trained backbone. Namely, we are given the extracted features of 1) a generic base class dataset $\mathcal{D}_{base} = \{\bm{x}^{base}_i\}_{i=1}^{N_{base}}\in\mathcal{C}_{base}$ that contains $N_{base}$ labelled samples where each sample $\bm{x}^{base}_i$ is a column vector of length $D$, and $\mathcal{C}_{base}$ is the set of base classes to which these samples belong. These base classes have been used to train the backbone. And similarly, 2) a novel class dataset $\mathcal{D}_{novel} = \{\bm{x}^{novel}_n\}_{n=1}^{N}$ containing $N$ samples belonging to a set of $K$ novel classes $\mathcal{C}_{novel}$ ($\mathcal{C}_{base} \cap \mathcal{C}_{novel} = \emptyset$). On this novel dataset, only a few elements are labelled, and the aim is to predict the missing labels. Denote $\mathbf{X}$ the matrix obtained by aggregating elements in $\mathcal{D}_{novel}$ row-wise.

When benchmarking transductive FSC methods, it is common to randomly generate Few-Shot tasks by sampling $\mathcal{D}_{novel}$ from a larger dataset. These tasks are generated by sampling $K$ distinct classes, $L$ distinct labelled elements for each class (called support set) and $Q$ total unlabelled elements without repetition and distinct from the labelled ones (called query set). All these unlabelled elements belong to one of the selected classes. We obtain a total of $N = KL + Q$ elements in the task, and compute the accuracy on the $Q$ unlabelled ones. Depending on how unlabelled instances are distributed among selected classes within a task, we further distinguish a balanced setting where the query set is evenly distributed among the $K$ classes, from an unbalanced setting where it can vary from class to class. An automatic way to generate such unbalanced Few-Shot tasks has been proposed in~\cite{veilleux2021realistic} where the number of elements to draw from each class is determined using a Dirichlet distribution parameterized by $\alpha_o^{*}\mathbf{1}$, where $\mathbf{1}$ is the all-one vector. To solve a transductive FSC task, our method is composed of PLDA and VB inference, that we introduce in the next paragraphs.

\subsection{Probabilistic Linear Discriminant Analysis (PLDA)}
\label{plda}
In our work, PLDA~\cite{ioffe2006probabilistic} is mainly used to reduce feature dimensions. For a Few-Shot task $\mathbf{X}$, let $\bm{\Phi}_w$ be a positive definite matrix representing the estimated shared within-class covariance of a given class, and $\bm{\Phi}_b$ be a positive semi-definite matrix representing the estimated between-class covariance that generates class variables. The goal of PLDA is to project data onto a subspace while maximizing the signal-to-noise ratio for class labelling. In details, we obtain a projection matrix $\mathbf{W}$ that diagonalizes both $\bm{\Phi}_w$ and $\bm{\Phi}_b$ and yield the following equations:
\begin{equation}
    \begin{aligned}
    \mathbf{W}^{T}\bm{\Phi}_w\mathbf{W} &= \mathbf{I},\quad \mathbf{W}^{T}\bm{\Phi}_b\mathbf{W} = \bm{\Psi} 
    \end{aligned}
\label{eq: diagonalization}
\end{equation}

where $\mathbf{I}$ is an identity matrix and $\bm{\Psi}$ is a diagonal matrix. In this paper, we assume a similar distribution between the pre-trained base classes and the transferred novel classes~\cite{DBLP:conf/iclr/YangLX21}. Therefore we propose to estimate $\bm{\Phi}_w$ to be the within-class scatter matrix of $\mathcal{D}_{base}$, denoted as $\mathbf{S}_{w}^{bass}$. In practice we implement PLDA by firstly transforming $\mathbf{X}$ using a rotation matrix $\mathbf{R}\in\mathbb{R}^{D\times D}$ and a set of scaling values $\bm{s}\in\mathbb{R}^D$ obtained from $\mathbf{S}_{w}^{base}$. Note that we clamp the scaling values to be no larger than an upper-bound $s_{max}$ in order to prevent too large values, $s_{max}$ is a hyper-parameter. Then we project the transformed data onto their estimated class centroids space, in accordance with the $d$ largest eigenvalues of $\bm{\Psi}$, and obtain dimension-reduced data $\mathbf{U}=[\bm{u}_1,...,\bm{u}_n,...,\bm{u}_N]^T\in\mathbb{R}^{N\times d}$ where $\bm{u}_n = \mathbf{W}^T\bm{x}_n$ and $d=K-1$. More detailed implementation can be found in Appendix.

\subsection{Variational Bayesian (VB) Inference}
\label{vb}

During VB inference, we operate on a reduced $d$-dimensional space obtained after applying PLDA. Considering a Gaussian mixture model for a given task $\mathbf{U}\in\mathbb{R}^{N\times d}$ in reduced space, let $\theta$ be the unknown variables of the model. In VB we attempt to find a probability distribution $q(\theta)$ that approximates the true posterior $p(\theta|\mathbf{U})$, i.e. maximizes the ELBO (see Appendix for more details). In our case, we define $\theta = \{\bm{Z}, \bm{\pi}, \bm{\mu}\}$ where $\bm{Z} = \{\bm{z}_n\}_{n=1}^N$ is a set of latent variables used as class indicators, each latent variable $\bm{z}_n$ has an one-of-K representation, $\bm{\pi}$ is a $K$-dimensional vector representing mixing ratios between the classes, and $\bm{\mu}=\{\bm{\mu}_k\}_{k=1}^{K}$ where $\bm{\mu}_k$ is the centroid for class $k$. Note that 1) contrary to EM where $\bm{\pi}, \bm{\mu}$ are seen as parameters that can be estimated directly, in VB they are deemed as hidden variables following certain distribution laws. 2) This is not a full VB model due to the lack of precision matrix (i.e. the inverse of covariance matrix) as a variable in $\theta$. Although a VB model frees up more parameters for the unknown variables, it also increases the instability in estimations so that the model becomes too sensible. Therefore, in this paper we impose an assumption that all classes in $\mathbf{U}$ share the same precision matrix and it is fixed during VB iterations. Namely we define $\bm{\Lambda}_k=\bm{\Lambda}=T_{vb} \mathbf{I}$ for $k=1,...,K$, where $T_{vb}$ is a hyper-parameter in order to compensate the variation between base and estimated novel class distributions.

In order for a model to be in a variational bayesian setting, we define priors and likelihoods on the unknown variables, with several initialization parameters attached:
\begin{equation}
    \begin{aligned}
    \text{priors}: \quad
    p(\bm{\pi}) &= Dir(\bm{\pi}|\alpha_o), \quad
    p(\bm{\mu}) = \prod_{k=1}^K \mathcal{N}(\bm{\mu}_k|\bm{m}_o, (\beta_o\bm{\Lambda})^{-1}), \\
    \text{likelihoods}:\quad
    p(\bm{Z}|\bm{\pi}) &= \prod_{n=1}^N Categorical(\bm{z}_n|\bm{\pi}), \quad
    p(\bm{U}|\bm{Z}, \bm{\mu}) = \prod_{n=1}^N \prod_{k=1}^K \mathcal{N}(\bm{u}_n|\bm{\mu}_k, \bm{\Lambda}^{-1})^{z_{nk}}
    \end{aligned}
\label{eq: prior}
\end{equation}

where $\bm{\pi}$ follows a K-dimensional symmetric Dirichlet distribution, with $\alpha_o$ being the prior of component weight for each class, which we set to $2.0$ in accordance with~\cite{veilleux2021realistic}, i.e. the same value as the Dirichlet distribution parameter $\alpha_o^{*}$ that is used to generate Few-Shot tasks. The vector $\bm{m}_o$ is the prior about the class centroid variables, we let it to be $\mathbf{0}$. And $\beta_o$ stands for the prior about the moving range of class centroid variables: the larger it is, the closer the centroids are to $\bm{m}_o$. We empirically found that $\beta_o=10.0$ gives consistent good results across datasets and FSC problems.

As previously stated, we approximate a variable distribution to the true posterior. To further simplify, we follow the Mean-Field assumption~\cite{prezhdo1999mean, jaakkola1998improving} and assume that the unknown variables are independent from one another. Therefore we let $q(\theta)=q(\bm{Z}, \bm{\pi}, \bm{\mu})=q(\bm{Z})q(\bm{\pi})q(\bm{\mu}) \approx p(\bm{Z}, \bm{\pi}, \bm{\mu}/\mathbf{U})$ and solve for each term. The explicit formulation for these marginals is provided in Eq.~\ref{eq: m_step_1},~\ref{eq: e_step_1} (see Appendix for more details). The estimation of the various parameters is then classically performed through an iterative EM framework as presented further.  

Denote $\bm{o}_n=[o_{n1}, ..., o_{nk},..., o_{nK}]$ as the soft class assignment for $\bm{u}_n$ ($o_{nk}\ge 0,\ \sum_{k=1}^K o_{nk} = 1)$, and $o_{nk}$ represents the portion of $n$th sample allocated to $k$th class.

\paragraph{M step:} In this step we estimate $q(\bm{\pi})$ and $q(\bm{\mu})$ in use of the class assignments $o_{nk}$:

\begin{equation}
    \begin{aligned}
    p(\bm{\pi}) = Dir(\bm{\pi}|\alpha_o)\ &\implies\  
    q^*(\bm{\pi}) = Dir(\bm{\pi}|\bm{\alpha}) \quad\text{with}\quad \alpha_k = \alpha_o + N_k, \\
    p(\bm{\mu}) = \prod_{k=1}^K \mathcal{N}(\bm{\mu}_k|\bm{m}_o, (\beta_o\bm{\Lambda})^{-1})
     \ &\implies\ q^*(\bm{\mu}) = \prod_{k=1}^K\mathcal{N}(\bm{\mu}_k|\bm{m}_k, (\beta_k\bm{\Lambda})^{-1}) \\
     \quad&\text{with}\quad \beta_k = \beta_o + N_k,\ \bm{m}_k = \frac{1}{\beta_k}(\beta_o\bm{m}_o + \sum_{n=1}^N o_{nk}\bm{u}_n),
    \end{aligned}
\label{eq: m_step_1}
\end{equation}

where $\bm{\alpha} = [\alpha_1,...,\alpha_k,...,\alpha_K]$ are the estimated component weights for classes, and $N_k = \sum_{n=1}^N o_{nk}$ is the sum of the soft assignments for all samples in class $k$. We also estimate the moving range parameter $\beta_k$ and the centroid $\bm{m}_k$ for each class centroid variable. We observe that the posteriors take the same forms as the priors. Demonstration of these results is presented in Appendix.

\paragraph{E step:} In this step we estimate $q(\bm{Z})$ by updating $o_{nk}$, using the current values of all other parameters computed in the M-step, i.e. $\alpha_k$, $\beta_k$ and $\bm{m}_k$.

\begin{equation}
    \begin{aligned}
    p(\bm{Z}|\bm{\pi}) &= \prod_{n=1}^N Categorical(\bm{z}_n|\bm{\pi})\ \implies\ q^*(\bm{Z}) = \prod_{n=1}^N Categorical(\bm{z}_n|\bm{o}_n)
    \end{aligned}
\label{eq: e_step_1}
\end{equation}

where each element $o_{nk}$ can be computed as $o_{nk} = \frac{\rho_{nk}}{\sum_{j=1}^K \rho_{nj}}$ in which:  
\begin{equation}
    \begin{aligned}
    \log\rho_{nk} = \psi(\alpha_k)-\psi(\sum_{j=1}^K \alpha_j) 
    + \frac{1}{2}\log|\bm{\Lambda}| - \frac{d}{2}\log 2\pi - \frac{1}{2}[d\beta_k^{-1}+(\bm{u}_n-\bm{m}_k)^{T}\bm{\Lambda}(\bm{u}_n-\bm{m}_k)],
    \end{aligned}
\label{eq: e_step_2}
\end{equation}

with $\psi(\cdot)$ being the logarithmic derivative of the gamma function (also known as the digamma function). We observe that $q^*(\bm{Z})$ follows the same categorical distribution as the likelihood, and it is parameterized by $o_{nk}$. More details can be found in Appendix.

\paragraph{Proposed algorithm} The proposed method combines PLDA and VB inference which leads to an Efficiency Guided Adaptive Dimension Reduction for VAriational BAyesian inference. We thus name our proposed method ``BAVARDAGE'', and the detailed description is presented in Algorithm~\ref{alg:bavardage}. Given a Few-Shot task $\mathbf{X}$ and a within-class scatter matrix $\mathbf{S}_w^{base}$, we initialize $o_{nk}$ using EM algorithm with an assumed covariance matrix, adjusted by a temperature hyper-parameter $T_{km}$, for all classes. Note that this is equivalent to Soft-KMEANS~\cite{kearns1998information} algorithm. And for each iteration we update parameters: in M step we update $\alpha_k, \beta_k$ and centroids $\bm{m}_k$, in E step we only update $o_{nk}$, and we apply PLDA with the updated $o_{nk}$ to reduce feature dimensions. Finally, predicted labels are obtained by selecting the class that corresponds to the largest value in $o_{nk}$.

The illustration of our proposed method is presented in Figure~\ref{fig:figure}. For a Few-Shot task that has three classes (red, blue and green) with unlabelled samples depicted on the probability simplex, we firstly initialize $o_{nk}$ with Soft-KMEANS which directs some data points to their belonging classes while further distancing some points from their targeted classes. Then we apply the proposed VB inference integrated with PLDA, resulting in additional points moving towards their corresponding classes.  

\begin{algorithm}[tb]
   \caption{BAVARDAGE}
   \label{alg:bavardage}
\begin{algorithmic}
   \STATE {\bfseries Inputs:} {$\mathbf{X}\in\mathbb{R}^{N\times D},\ \mathbf{S}_w^{base}\in\mathbb{R}^{D\times D}$}
   \STATE {\bfseries Hyper-parameters:} {$T_{km},\ T_{vb},\ s_{max}$}
   \STATE {\bfseries Priors for VB:} {$\alpha_o=2.0,\, \beta_o=10.0,\, \bm{m}_o=0,\, \bm{\Lambda} = T_{vb}\cdot\mathbf{I}$}
   \STATE {\bfseries Initializations:} {$o_{nk}=\text{EM}\ (\mathbf{X},\, T_{km})$}
   \FOR{$i=1$ {\bfseries to} $n_{step}$}
   \STATE $\mathbf{U}= \text{PLDA}\ (\mathbf{X},\ \mathbf{S}_w^{base},\ s_{max},\ o_{nk})$ \quad \# See more details in Appendix. 
   \STATE {\bfseries VB (M step)}:
   \STATE $\alpha_k=\alpha_o+\sum_{n=1}^N o_{nk}$
   \STATE $\beta_k=\beta_o+\sum_{n=1}^N o_{nk}$
   \STATE $\bm{m}_k=\frac{1}{\beta_k}(\beta_o\bm{m}_o+\sum_{n=1}^N o_{nk}\bm{u}_n)$
   \STATE {\bfseries VB (E step)}:
   \STATE $\log{\rho_{nk}}=\psi(\alpha_k)-\psi(\sum_{j=1}^K \alpha_j) + \frac{1}{2}\log|\bm{\Lambda}| - \frac{d}{2}\log 2\pi - \frac{1}{2}[d\beta_k^{-1}+(\bm{u}_n-\bm{m}_k)^{T}\bm{\Lambda}(\bm{u}_n-\bm{m}_k)]$
   \STATE $o_{nk} = \frac{\rho_{nk}}{\sum_{j=1}^K \rho_{nj}}$
   \ENDFOR
\STATE {\bfseries return} $\hat{\ell}(\bm{x}_n)=\arg\max_k(o_{nk})$
\end{algorithmic}
\end{algorithm}

\section{Experiments}
\label{experiments}

In this section we provide details on the standard transductive Few-Shot classification settings, and we evaluate the performance of our proposed method. 

\paragraph{Benchmarks} We test our method on standard Few-Shot benchmarks: \textit{mini}-Imagenet~\cite{russakovsky2015imagenet}, \textit{tiered}-Imagenet~\cite{DBLP:conf/iclr/RenTRSSTLZ18} and caltech-ucsd birds-200-2011 (CUB)~\cite{wah2011caltech}. \textit{mini}-Imagenet is a subset of ILSVRC-12~\cite{russakovsky2015imagenet} dataset, it contains a total of $60,000$ images of size $84\times 84$ belonging to $100$ classes ($600$ images per class) and follows a 64-16-20 split for base, validation and novel classes. \textit{tiered}-Imagenet is a larger subset of ILSVRC-12 containing $608$ classes with $779,165$ images of size $84\times 84$ in total, and we use the standard 351-97-160 split, and CUB is composed of $200$ classes following a 100-50-50 split (Image size: $84\times 84$). In Appendix we also show the performance of our proposed method on other well-known Few-Shot benchmarks such as FC100~\cite{oreshkin2018tadam} and CIFAR-FS~\cite{DBLP:conf/iclr/BertinettoHTV19}.

\paragraph{Settings} Following previous works~\cite{lichtenstein2020tafssl, rodriguez2020embedding, veilleux2021realistic}, our proposed method is evaluated on $1$-shot $5$-way ($K=5$, $L=1$), and $5$-shot $5$-way ($K=5$, $L=5$) scenarios. As for the query set, we set a total number of $Q=75$ unlabelled samples, from which we further define two  settings: 1) a balanced setting where unlabelled instances are evenly distributed among $K$ classes, and 2) an unbalanced setting where the query set is randomly distributed, following a Dirichlet distribution parameterized by $\alpha_o^{*}$. In our paper we follow the same setting as~\cite{veilleux2021realistic} and set $\alpha_o^{*}=2.0$, further experiments with different values are conducted in the next sections. The performance of our proposed method is evaluated by computing the averaged accuracy over $10,000$ Few-Shot tasks.  

\paragraph{Implementation details} In this paper we firstly compare our proposed algorithm with the other state-of-the-art methods using the same pretrained backbones and benchmarks provided in~\cite{veilleux2021realistic}.  Namely we extract the features using the same ResNet-18 (RN18) and WideResNet28\_10 (WRN) neural models, and present the performance on \textit{mini}-Imagenet, \textit{tiered}-Imagenet and CUB datasets. In our proposed method, the raw features are preprocessed following~\cite{DBLP:journals/corr/abs-1911-04623}. As for the hyper-parameters, we set $T_{km}=10, T_{vb}=50, s_{max}=2$ for the balanced setting; $T_{km}=50, T_{vb}=50, s_{max}=1$ for the unbalanced setting, and we use the same VB priors for all settings. To further show the functionality of our proposed method on different backbones and other benchmarks, we tested BAVARDAGE on a recent high performing feature extractor trained on a ResNet-12 (RN12) neural model~\cite{mangla2020charting, bendou2022easy}, and we report the accuracy in Table~\ref{results_1} and in Appendix with various settings.        

\subsection{Main results} 
\label{results}

The main results on the relevant settings are presented in Table~\ref{results_1}. Note that we report the accuracy of other methods following~\cite{veilleux2021realistic}, and add the performance of our proposed method in comparison, using the same pretrained RN18 and WRN feature extractors, and we also report the result of a RN12 backbone pretrained following~\cite{bendou2022easy}. We observe that our proposed algorithm reaches state-of-the-art performance for nearly all referenced datasets in the unbalanced setting, surpassing previous methods by a noticeable margin especially on 1-shot. In the balanced setting we also reach competitive accuracy compared with~\cite{hu2021leveraging} along with other works that make use of a perfectly balanced prior on unlabelled samples, while our proposed method suggests no such prior. In addition, we provide results on the other Few-Shot benchmarks with different settings in Appendix. 

\begin{table}[h]
  \caption{Comparisons of the state-of-the-art methods on \textit{mini}-Imagenet, \textit{tiered}-Imagenet and CUB datasets using the same pretrained backbones as~\cite{veilleux2021realistic}, along with the accuracy of our proposed method on a ResNet-12 backbone pretrained following~\cite{bendou2022easy}.}
  \label{results_1}
  \centering
  \resizebox{\columnwidth}{!}{
  \begin{tabular}{lccccc}
    \toprule
    \textbf{\textit{mini}-Imagenet} & & \multicolumn{2}{c}{\textbf{unbalanced}} & \multicolumn{2}{c}{\textbf{balanced}} \\
    \cmidrule(r){3-6}
    Method & Backbone & \textbf{1-shot} & \textbf{5-shot} & \textbf{1-shot} & \textbf{5-shot} \\
    \midrule
    MAML~\cite{finn2017model} & \multirow{9}{*}{\small RN18/WRN~\cite{veilleux2021realistic}}  & $47.6/-$ & $64.5/-$ & $51.4/-$ & $69.5/-$    \\
    Versa~\cite{DBLP:conf/iclr/GordonBBNT19} &  & $47.8/-$ & $61.9/-$ & $50.0/-$ & $65.6/-$  \\
    Entropy-min~\cite{DBLP:conf/iclr/DhillonCRS20} &  & $58.5/60.4$ & $74.8/76.2$ & $63.6/66.1$ & $82.1/84.2$ \\
    PT-MAP~\cite{hu2021leveraging} &  & $60.1/60.6$ & $67.1/66.8$ & $\mathbf{76.9}/\mathbf{78.9}$ & $\mathbf{85.3}/86.6$ \\
    LaplacianShot~\cite{ziko2020laplacian} & & $65.4/70.0$ & $81.6/83.2$ & $70.1/72.9$ & $82.1/83.8$ \\
    BD-CSPN~\cite{liu2020prototype} & & $67.0/70.4$ & $80.2/82.3$ & $69.4/72.5$ & $82.0/83.7$ \\
    TIM~\cite{NEURIPS2020_196f5641} & & $67.3/69.8$ & $79.8/81.6$ & $71.8/74.6$ & $83.9/85.9$ \\
    $\alpha$-TIM~\cite{veilleux2021realistic} & & $67.4/69.8$ & $82.5/84.8$ & $-/-$ & $-/-$ \\
    \rowcolor{LightCyan}
    BAVARDAGE (ours) & & $\mathbf{71.0}/\mathbf{74.1}$ & $\mathbf{83.6}/\mathbf{85.5}$ & $75.1/78.5$ & $84.5/\mathbf{87.4}$\\
    \midrule
    \rowcolor{LightCyan}
    BAVARDAGE (ours) & RN12~\cite{bendou2022easy} & $\mathbf{77.8}$ &  $\mathbf{88.0}$ & $\mathbf{82.7}$ & $\mathbf{89.5}$\\
    \bottomrule
    
    \textbf{\textit{tiered}-Imagenet} & & \multicolumn{2}{c}{\textbf{unbalanced}} & \multicolumn{2}{c}{\textbf{balanced}} \\
    \cmidrule(r){3-6}
    Method & Backbone & \textbf{1-shot} & \textbf{5-shot} & \textbf{1-shot} & \textbf{5-shot} \\
    \midrule
    Entropy-min~\cite{DBLP:conf/iclr/DhillonCRS20} & \multirow{7}{*}{\small RN18/WRN~\cite{veilleux2021realistic}}  & $61.2/62.9$ & $75.5/77.3$ & $67.0/68.9$ & $83.1/84.8$   \\
    PT-MAP~\cite{hu2021leveraging} &  & $64.1/65.1$ & $70.0/71.0$ & $\mathbf{82.9}/\mathbf{84.6}$ & $\mathbf{88.8}/\mathbf{90.0}$ \\
    LaplacianShot~\cite{ziko2020laplacian} & & $72.3/73.5$ & $85.7/86.8$ & $77.1/78.8$ & $86.2/87.3$ \\
    BD-CSPN~\cite{liu2020prototype} &  & $74.1/75.4$ & $84.8/85.9$ & $76.3/77.7$ & $86.2/87.4$ \\
    TIM~\cite{NEURIPS2020_196f5641} &  & $74.1/75.8$ & $84.1/85.4$ & $78.6/80.3$ & $87.7/88.9$\\
    $\alpha$-TIM~\cite{veilleux2021realistic} &  & $74.4/76.0$ & $\mathbf{86.6}/\mathbf{87.8}$ & $-/-$ & $-/-$ \\
    \rowcolor{LightCyan}
    BAVARDAGE (ours) & & $\mathbf{76.6}/\mathbf{77.5}$ & $86.5/87.5$ & $80.3/81.5$ & $87.1/88.3$\\
    \midrule
    \rowcolor{LightCyan}
    BAVARDAGE (ours) & RN12~\cite{bendou2022easy} & $\mathbf{79.4}$ &  $\mathbf{88.0}$ & $\mathbf{83.5}$ & $\mathbf{89.0}$\\
    \bottomrule
    
    \textbf{CUB} & & \multicolumn{2}{c}{\textbf{unbalanced}} & \multicolumn{2}{c}{\textbf{balanced}} \\
    \cmidrule(r){3-6}
    Method & Backbone & \textbf{1-shot} & \textbf{5-shot} & \textbf{1-shot} & \textbf{5-shot} \\
    \midrule
    PT-MAP~\cite{hu2021leveraging} & \multirow{7}{*}{RN18~\cite{veilleux2021realistic}} & $65.1$ & $71.3$ & $85.5$ & $91.3$   \\
    Entropy-min~\cite{DBLP:conf/iclr/DhillonCRS20} &  & $67.5$ & $82.9$ & $72.8$ & $88.9$ \\
    LaplacianShot~\cite{ziko2020laplacian} &  & $73.7$ & $87.7$ & $78.9$ & $88.8$ \\
    BD-CSPN~\cite{liu2020prototype} &  & $74.5$ & $87.1$ & $77.9$ & $88.9$ \\
    TIM~\cite{NEURIPS2020_196f5641} &  & $74.8$ & $86.9$ & $80.3$ & $90.5$ \\
    $\alpha$-TIM~\cite{veilleux2021realistic} &  & $75.7$ & $89.8$ & $-$ & $-$ \\
    \rowcolor{LightCyan}
    BAVARDAGE (ours) & & $\mathbf{82.0}$ & $\mathbf{90.7}$ & $\mathbf{85.6}$ & $\mathbf{91.4}$ \\
    \midrule
    \rowcolor{LightCyan}
    BAVARDAGE (ours) & RN12~\cite{bendou2022easy} & $\mathbf{83.1}$ &  $\mathbf{90.8}$ & $\mathbf{87.4}$ & $\mathbf{92.0}$\\
    \bottomrule
  \end{tabular}
  }
\end{table}

\subsection{Ablation studies} 
\label{ablation}

\paragraph{Analysis on the elements of BAVARDAGE} In this experiment we dive into our proposed method and conduct an ablation study on the impact of each element. Namely, we report the performance in the following 3 scenarios: 1) only run Soft-KMEANS on the extracted features to obtain a baseline accuracy; 2) run the VB model with $o_{nk}$ initialized by Soft-KMEANS, without reducing the feature space; and 3) integrate PLDA into VB iterations. From Table~\ref{tab:ablation} we observe only a slight increase of accuracy compared with baseline when no dimensionality reduction is applied. This is due to the fact that high feature dimensions increase uncertainty in the estimations, making the model sensitive to parameters. With our implementation of PLDA iteratively applied in the VB model, we can see from the table that the performance increases by a relatively large margin, suggesting the effectiveness of our proposed adaptive dimension reduction method.     

\begin{table}[h]
    \caption{Ablation study on the elements of our proposed method, with results tested on \textit{mini}-Imagenet (backbone: WRN) and CUB (backbone: RN18) in the unbalanced setting.}
    \label{tab:ablation}
    \centering
    \begin{tabular}{ccccccc}
         \toprule
         & & & \multicolumn{2}{c}{\textbf{\textit{mini}-Imagenet}} & \multicolumn{2}{c}{\textbf{CUB}} \\
         \cmidrule(r){4-7}
         Soft-KMEANS & VB & PLDA & \textbf{1-shot} & \textbf{5-shot} & \textbf{1-shot} & \textbf{5-shot} \\
         
         \midrule
         \checkmark &  & & $71.4$& $82.4$ & $77.5$ & $86.7$\\
         
         \midrule
         \checkmark&\checkmark & & $71.8$& $82.5$ & $77.8$ &$87.2$\\
         
         \midrule
         \checkmark &\checkmark &\checkmark & $\mathbf{74.1}$ & $\mathbf{85.5}$& $\mathbf{82.0}$&$\mathbf{90.7}$\\
         
         \bottomrule
         
    \end{tabular}
\end{table}

\begin{figure}[h]
\centering
\subfloat[T-SNE]{\includegraphics[width=0.25\linewidth]{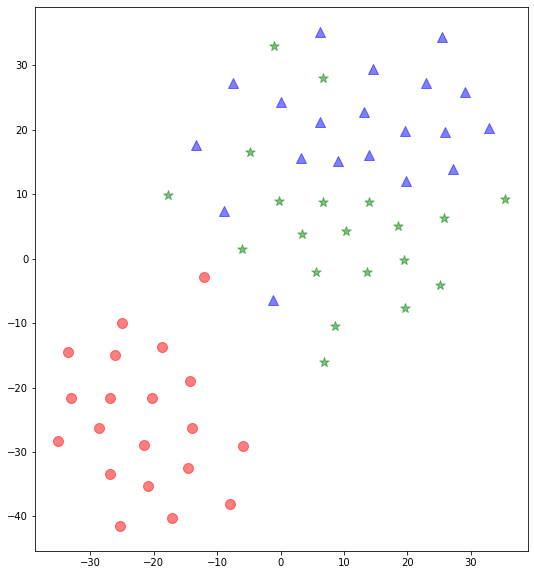}} 
\subfloat[Centroids projection]{\includegraphics[width=0.25\linewidth]{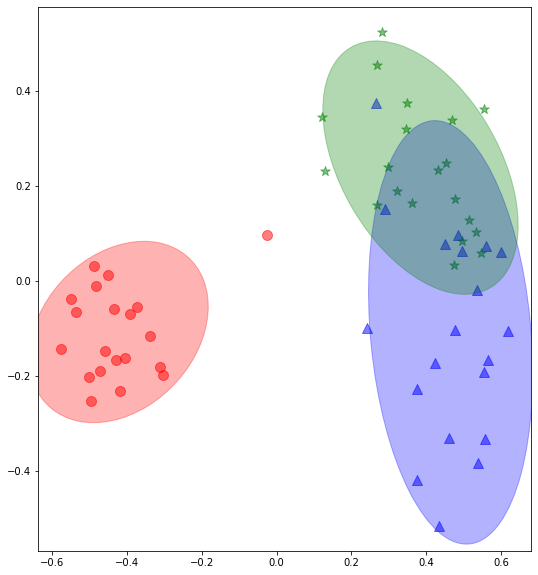}}
\subfloat[PLDA]{\includegraphics[width=0.25\linewidth]{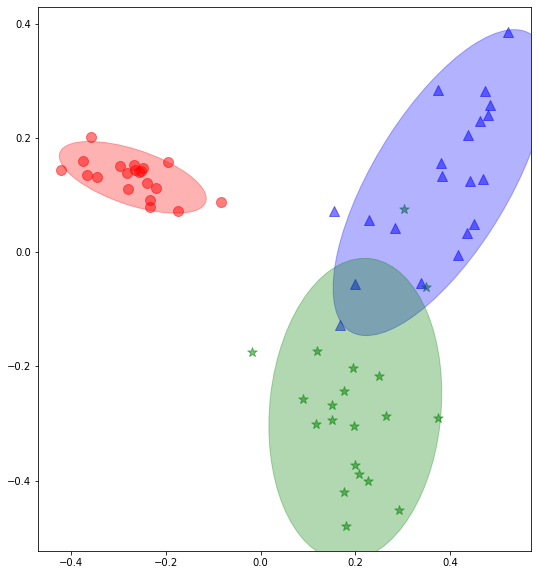}}
\caption{Visualization of extracted features of a Few-Shot task using different projection methods (dataset: \textit{mini}-Imagenet, backbone: WRN), we report a $86.7\%$, $90.0\%$ and $95.0\%$ prediction accuracy corresponding to each projection.}
\label{fig:projection}
\end{figure}

\paragraph{Visualization of features for different projections} To further showcase the effect of proposed PLDA, in Fig.~\ref{fig:projection} we visualize the extracted features of a 3-way Few-Shot task in the following 3 scenarios: (a) features in the original space, using T-SNE~\cite{van2008visualizing} for visualization purpose; (b) features that are projected directly onto their centroids space, and finally (c) features projected using PLDA. The ellipses drawn in (b) and (c) are the cluster estimations computed using the real labels of data samples, and we can thus observe a larger separation of different clusters with PLDA projection for the task in which the original features overlap heavily between clusters in blue and green.  

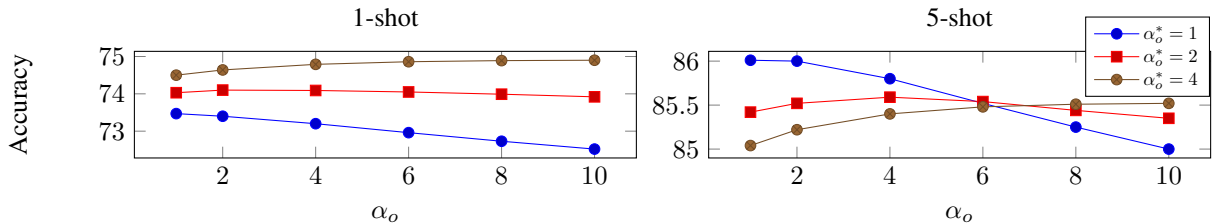
\begin{figure}[H]
    \begin{tikzpicture}
       \begin{scope}[]
        \begin{axis}[
            title=1-shot,
            x label style={at={(axis description cs:0.5,0.0)},anchor=north},
            y label style={at={(axis description cs:0.0,.5)},rotate=0,anchor=south},
            xlabel=$\alpha_o$,
            ylabel=Accuracy,
            height=3.0cm,
            width=.5\textwidth,
            ]
          
          \addplot coordinates
          {(1.0, 73.47) (2.0, 73.40) (4.0, 73.20) (6.0, 72.96) (8.0, 72.73) (10.0, 72.52)};
          
          \addplot coordinates
          {(1.0, 74.03) (2.0, 74.10) (4.0, 74.09) (6.0, 74.05) (8.0, 73.99) (10.0, 73.92)};
          
          \addplot coordinates
          {(1.0, 74.50) (2.0, 74.64) (4.0, 74.79) (6.0, 74.86) (8.0, 74.89) (10.0, 74.90)};
            
        \end{axis}
      \end{scope}
    \end{tikzpicture}
    \begin{tikzpicture}
       \begin{scope}[]
        \begin{axis}[
            title=5-shot,
            x label style={at={(axis description cs:0.5,0.0)},anchor=north},
            y label style={at={(axis description cs:0.0,.5)},rotate=0,anchor=south},
            xlabel=$\alpha_o$,
            height=3.0cm,
            width=.5\textwidth,
            legend style={nodes={scale=0.7, transform shape}, at={(0.75, 0.95)}, anchor=west},
            ]
          
          \addlegendentry{$\alpha_o^{*}=1$}
          \addplot coordinates
          {(1.0, 86.01) (2.0, 86.00) (4.0, 85.80) (6.0, 85.52) (8.0, 85.25) (10.0, 85.00)};
          
          \addlegendentry{$\alpha_o^{*}=2$}
          \addplot coordinates
          {(1.0, 85.42) (2.0, 85.52) (4.0, 85.59) (6.0, 85.54) (8.0, 85.44) (10.0, 85.35)};
          
          \addlegendentry{$\alpha_o^{*}=4$}
          \addplot coordinates
          {(1.0, 85.04) (2.0, 85.22) (4.0, 85.40) (6.0, 85.48) (8.0, 85.51) (10.0, 85.52)};
            
        \end{axis}
      \end{scope}
    \end{tikzpicture}
  \caption{1-shot and 5-shot accuracy on different imbalance levels (varying $\alpha_o^{*}$) as a function of VB priors $\alpha_o$ (dataset: \textit{mini}-Imagenet, backbone: WRN).}
  \label{fig:alpha}
\end{figure}

\paragraph{Robustness against imbalance} In Table~\ref{results_1} we show the accuracy of our proposed method using VB priors introduced in Section~\ref{vb}, in which $\alpha_o$ is set to be equal to the Dirichlet's parameter $\alpha_o^{*}$ for the level of imbalance in the query set. Therefore, in this experiment we test the robustness of BAVARDAGE, namely in Fig.~\ref{fig:alpha} we alter $\alpha_o$ and report the accuracy on different imbalance levels (varying $\alpha_o^{*}$) in both 1-shot and 5-shot settings. Note that the proposed model becomes slightly more sensitive to $\alpha_o$ when the level of imbalance increases (smaller $\alpha_o^{*}$), with an approximate $1\%$ drop of accuracy when increasing $\alpha_o$ in the case of $\alpha_o^{*}=1$.  

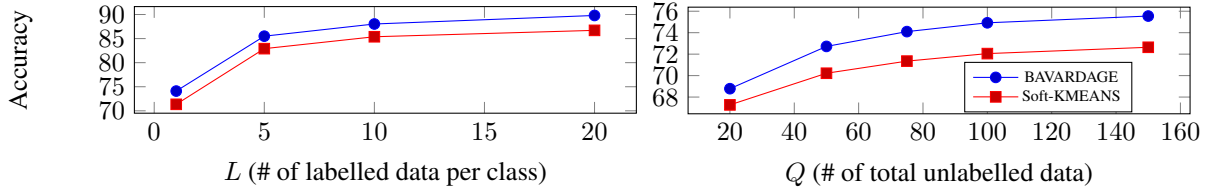
\begin{figure}[H]
    \begin{tikzpicture}
       \begin{scope}[]
        \begin{axis}[
            x label style={at={(axis description cs:0.5,0.0)},anchor=north},
            y label style={at={(axis description cs:0.0,.5)},rotate=0,anchor=south},
            xlabel=$L$ (\# of labelled data per class),
            ylabel=Accuracy,
            height=3.0cm,
            width=.5\textwidth,
            ]
          
          
          \addplot coordinates
          {(1, 74.1) (5, 85.5) (10, 88.03) (20, 89.8)};
          
          \addplot coordinates
          {(1, 71.35) (5, 82.91) (10, 85.38) (20, 86.71)};
          
        \end{axis}
      \end{scope}
    \end{tikzpicture}
    \begin{tikzpicture}
       \begin{scope}[]
        \begin{axis}[
            x label style={at={(axis description cs:0.5,0.0)},anchor=north},
            y label style={at={(axis description cs:0.0,.5)},rotate=0,anchor=south},
            xlabel=$Q$ (\# of total unlabelled data),
            height=3.0cm,
            width=.5\textwidth,
            legend style={nodes={scale=0.6, transform shape}, at={(0.55, 0.25)}, anchor=west},
            ]
          
          \addlegendentry{BAVARDAGE}
          \addplot coordinates
          {(20, 68.78) (50, 72.72) (75, 74.1) (100, 74.92) (150, 75.55)};
          
          \addlegendentry{Soft-KMEANS}
          \addplot coordinates
          {(20, 67.27) (50, 70.22) (75, 71.35) (100, 72.05) (150, 72.64)};
          
        \end{axis}
      \end{scope}
    \end{tikzpicture}
\caption{Accuracy as a funtion of $L$ and $Q$ in comparison with Soft-KMEANS (dataset: \textit{mini}-Imagenet, backbone: WRN).}
  \label{fig:vary}
\end{figure}

\paragraph{Varying Few-Shot settings} In this experiment we observe the performance of BAVARDAGE on different Few-Shot settings, namely we vary the number of labelled samples per class $L$ as well as the total number of unlabelled samples $Q$ in a task, for further comparison we also report the accuracy using only Soft-KMEANS algorithm. In Fig.~\ref{fig:vary} we can observe constant higher accuracy of our proposed method, and a slightly larger difference gap when $Q$ increases.  

\section{Conclusion}
\label{conclusion}

In this paper we proposed a clustering method based on Variational Bayesian Inference and Probabilistic Linear Discriminant Analysis for transductive Few-Shot Classification. BAVARDAGE has reached state-of-the-art accuracy on nearly all Few-Shot benchmarks in the realistic unbalanced setting, as well as competitive performance in the balanced setting without using a perfectly class-balanced prior. As our proposed method assumes a shared isotropic covariance matrix for all clusters, the estimations in VB models could be limited. Therefore the future work could study a better estimation of covariance matrices associated with each cluster. An interesting asset of the proposed method is that it performs most of its processing in a reduced $(K-1)$-dimensional space, where $K$ is the number of classes, suggesting interests for visualization and suitability for more elaborate statistical machine learning methods. As in~\cite{veilleux2021realistic}, we encourage the community to rethink the works in transductive settings to provide fairer grounds of comparison between the various proposed approaches.

\section{Appendix}
\label{appendix}

\subsection{Implementation details on the proposed PLDA}
\label{sup_plda}

In this section we present more details on our implementation of PLDA proposed in section 3.2 in the paper. Given $\mathbf{X}\in\mathbb{R}^{N\times D}$, we estimate its within-class covariance matrix to be $\mathbf{S}_w^{base}$ calculated from $\mathbf{D}_{base}$. Denote $\mathcal{I}^{base}_c$ as the set of samples belonging to base class $c$ where $c\in1,...,|\mathcal{C}_{base}|$, therefore $\bm{\Phi}_w$ is approximated as follows:

\begin{equation}
    \begin{aligned}
    \bm{\Phi}_w\approx \mathbf{S}_w^{base} = \frac{\sum_{c}\sum_{i\in\mathcal{I}^{base}_c}(\bm{x}^{base}_i-\bm{m}^{base}_c)(\bm{x}^{base}_i-\bm{m}^{base}_c)^T}{N_{base}}, 
    \end{aligned}
\label{eq: sup_sw_base}
\end{equation}

where $\bm{m}^{base}_c = \frac{1}{|\mathcal{I}^{base}_c|}\sum_{i\in\mathcal{I}^{base}_c} \bm{x}^{base}_i$ is the mean of $c$-th base class. Let $\bm{\lambda} = [\lambda_1,...,\lambda_i,...,\lambda_{D}]\in \mathbb{R}^{D}$ be the eigenvalues of $\mathbf{S}_w^{base}$ in descending order, and we set $\mathbf{R}=[\bm{r}_1,...,\bm{r}_i,...,\bm{r}_D]\in\mathbb{R}^{D\times D}$ to be the corresponding eigenvectors. In this paper we define a transformation matrix $\mathbf{T}=\mathbf{SR}$ where $\mathbf{S}$ is a diagonal matrix with diagonal values being the square root of multiplicative inverse of $\bm{\lambda}$, clamped to an upper bound $s_{max}$. Namely, $\bm{s}=diag(\mathbf{S})$ where $\bm{s}=[s_1,...s_i,...s_D]\in \mathbb{R}^{D}$ is a $D$-length vector containing the scaling value for each dimension, and we set $s_i$ to be as follows:

\begin{equation}
    \begin{aligned}
    s_i &= \left\{\begin{array}{ll}\lambda_i^{-0.5}& \text{if\quad} \lambda_i^{-0.5} \le s_{max}\\ s_{max} & \text{$otherwise$}\end{array}\right..
    \end{aligned}
\label{eq: sup_clamp}
\end{equation}

We can see from Eq.~\ref{eq: sup_clamp} that $\mathbf{T}$ is composed of a rotation matrix and scaling values on feature dimensions that help morph the within-class distribution into an identity covariance matrix. This corresponds to a data sphering/whitening process in which we decorrelate samples in each of the dimensions. In our implementation we transform $\mathbf{X}$ by multiplying it with $\mathbf{T}$. Therefore the sphered data samples, denoted as $\mathbf{X}'=[\bm{x}'_1,...\bm{x}'_n,...\bm{x}'_N]^T\in\mathbb{R}^{N\times D}$, are obtained from $\bm{x}'_n = \mathbf{T}\bm{x}_n$.

Next, we project $\mathbf{X}'$ onto a subspace that corresponds to the $K-1$ largest eigenvalues of its between-scatter matrix. Denote $\bm{m}'_k$ as the estimated centroid for class $k$, given soft class assignments $o_{nk}$ ($1\le n\le N, 1\le k \le K$), $\bm{m}'_k$ is computed as: 

\begin{equation}
    \begin{aligned}
    \bm{m}'_k &= \frac{\sum_{n=1}^N o_{nk}\bm{x}'_n}{\gamma + N_k},\ N_k = \sum_{n=1}^N o_{nk}\;,
    \end{aligned}
\label{eq: sup_mk}
\end{equation}
where $\gamma$ is used as an offset indicating how close the centroids are to $0$, in this paper we set it to $10.0$, same as $\beta_o$ in the VB model in reduced space. Therefore, the between-class scatter matrix $\bm{\Psi}$ of sphered samples can be calculated as:

\begin{equation}
    \begin{aligned}
    \bm{\Psi} = \sum_{k=1}^K (\bm{m}'_k - \bm{m}')(\bm{m}'_k - \bm{m}')^T\;,
    \end{aligned}
\label{eq: sup_psi}
\end{equation}
where $\bm{m}' = \frac{1}{K}\sum_{k=1}^K \bm{m}'_k$ is the mean of estimated class centroids. Then we project $\mathbf{X}'$ onto a $d$-length subspace, where $d=K-1$. In details, denote $\mathbf{V}=[\bm{v}_1,...,\bm{v}_i,...,\bm{v}_d]\in\mathbb{R}^{D\times d}$ to be the eigenvectors corresponding to the $d$ largest eigenvalues of $\bm{\Psi}$, the projected data $\mathbf{U}$ are obtained as $\bm{u}_n = \mathbf{V}^T\bm{x}'_n$ for each sample. Note that the formulation of $\bm{\Psi}$ in Eq.~\ref{eq: sup_psi} allows at most $K-1$ non-zero eigenvalues, therefore the resulting subspace projection using these eigenvectors is equivalent to a projection onto the affine subspace containing the centroids $\bm{m}'_k$. Furthermore, according to Eq. 1 in the paper, we can further deduce the projection matrix $\mathbf{W}$ to be as follows: 
\begin{equation}
    \begin{aligned}
    \bm{u}_n &= \mathbf{W}^T\bm{x}_n=\mathbf{V}^T\bm{x}'_n=\mathbf{V}^T\mathbf{T}\bm{x}_n=\mathbf{V}^T\mathbf{SR}\bm{x}_n, \\ 
    &\implies \mathbf{W} = (\mathbf{V}^T\mathbf{SR})^T = \mathbf{R}^T\mathbf{SV}.
    \end{aligned}
\label{eq: w}
\end{equation}

The entire process is described in Algorithm~\ref{alg:sup_plda}.

\begin{algorithm}[tb]
   \caption{Proposed PLDA}
   \label{alg:sup_plda}
\begin{algorithmic}
   \STATE {\bfseries Fonction} PLDA\ ($\mathbf{X},\ \mathbf{S}_w^{base},\ s_{max},\ o_{nk}$)
   \STATE \quad Sphere $\mathbf{X}$ using $\mathbf{T}$ (Eq.~\ref{eq: sup_clamp}), obtain $\mathbf{X}'$.
   \STATE \quad Estimate centroids $\bm{m}'_k$ using $o_{nk}$ (Eq.~\ref{eq: sup_mk}).
   \STATE \quad Compute $\bm{\Psi}$ using $\bm{m}'_k$ (Eq.~\ref{eq: sup_psi}).
   \STATE \quad Project $\mathbf{X}'$ onto the centroids space, obtain $\mathbf{U}$.
\STATE {\bfseries Return} $\mathbf{U}$
\end{algorithmic}
\end{algorithm}

\subsection{Implementation details on the proposed VB model}
\label{sup_vb}
In this section we provide more detailed explanation of our proposed VB model. Given a posterior $p(\theta|\mathbf{U})$, we approximate it with a function variational distribution $q(\theta)$ by minimizing the Kullback-Leibler divergence:
\begin{equation}
    \begin{aligned}
    q^*(\theta) &= \arg\min_{q} \{D_{KL}(q||p)\} \\
    & = \arg\min_{q} \{\log p(\mathbf{U})-\mathcal{L}(q)\} \\
    & = \arg\max_{q} \{\mathcal{L}(q)\}
    \end{aligned}
\label{eq: sup_kl}
\end{equation}

where the evidence $\log p(\mathbf{U})$ is considered fixed, and $\mathcal{L}(q)= \int q(\theta)\log\frac{p(\theta,\mathbf{U})}{q(\theta)}d\theta$ stands for Evidence Lower BOund (ELBO) providing ``evidence'' that we have chosen the right model. We can see that minimizing the Kullback-Leibler divergence is equivalent to maximizing the ELBO. Suppose $\theta = \{\theta_1,...,\theta_m,...,\theta_M\}$, we firstly factorize $q(\theta)=\prod_{m=1}^M q(\theta_m)$ according to the Mean-Field assumption, then we solve each term individually:
\begin{equation}
    \begin{aligned}
    \mathcal{L}(q) &= \int q(\theta)\log\frac{p(\theta,\mathbf{U})}{q(\theta)}d\theta \\
    &= \int \left(\prod_{m=1}^M q(\theta_m)\right)\left(\log p(\theta, \mathbf{U}) - \sum_{m=1}^M \log q(\theta_m)\right)d\theta_1 d\theta_2...d\theta_M \\
    &= \sum_{m=1}^M \left(\int q(\theta_m)\left(\int q(\theta_{-m})\log p(\theta,\mathbf{U})d\theta_{-m}\right)d\theta_m - \int q(\theta_m)\log q(\theta_m)d\theta_m\right),
    \end{aligned}
\label{eq: sup_q_1}
\end{equation}

and the ELBO is maximized when: 
\begin{equation}
    \begin{aligned}
    \log q^*(\theta_m)=\mathbb{E}_{\theta_{-m}}[\log p(\theta, \mathbf{U})]+const,
    \end{aligned}
\label{eq: sup_q_2}
\end{equation}

where $\mathbb{E}_{\theta_{-m}}[\cdot]$ stands for the expectation with respect to all variables in $\theta$ except $\theta_m$. In our method we define $\theta = \{\bm{Z}, \bm{\pi}, \bm{\mu}\}$, the detailed formula of some variables are presented as follows:
\begin{equation}
    \begin{aligned}
    \bm{z}_n &= [z_{n1}, ..., z_{nk},..., z_{nK}]\in\{0, 1\}^{K}, \quad \sum_{k=1}^K z_{nk} = 1, \\
    \bm{\pi} &= [\pi_1, ..., \pi_k, ..., \pi_K], \quad \pi_{k} \geq 0, \quad \sum_{k=1}^K \pi_{k} = 1.
    \end{aligned}
\label{eq: sup_latent}
\end{equation}

According to Bayes' theorem, we rewrite the posterior to be: 
\begin{equation}
    \begin{aligned}
    p(\theta|\mathbf{U}) = p(\bm{Z}, \bm{\pi}, \bm{\mu}|\mathbf{U}) &= \frac{p(\bm{Z}, \bm{\pi}, \bm{\mu}, \mathbf{U})}{p(\mathbf{U})} \\
    & = \frac{p(\mathbf{U}|\bm{Z}, \bm{\mu})p(\bm{Z}|\bm{\pi})p(\bm{\pi})p(\bm{\mu})}{p(\mathbf{U})},
    \end{aligned}
\label{eq: sup_bayes}
\end{equation}

in which:
\begin{equation}
    \begin{aligned}
    p(\mathbf{U}|\bm{Z}, \bm{\mu}) &= \prod_{n=1}^{N} \prod_{k=1}^{K} \mathcal{N}(\bm{u}_n|\bm{\mu}_k, \bm{\Lambda}^{-1})^{z_{nk}}, \\
    p(\bm{Z}|\bm{\pi}) &= \prod_{n=1}^N Categorical(\bm{z}_n|\bm{\pi}) = \prod_{n=1}^{N} \prod_{k=1}^{K} \pi_k^{z_{nk}}, \\
    p(\bm{\pi}) &= Dir(\bm{\pi}|\alpha_o) = \frac{\Gamma(\sum_{k=1}^K K\alpha_o)}{\prod_{k=1}^K \Gamma(\alpha_o)} \prod_{k=1}^K \pi_k^{\alpha_o-1} = C(\alpha_o)\prod_{k=1}^K \pi_k^{\alpha_o-1}, \\
    p(\bm{\mu}) &= \prod_{k=1}^K \mathcal{N}(\bm{\mu}_k|\bm{m}_o, (\beta_o\bm{\Lambda})^{-1}).
    \end{aligned}
\label{eq: sup_p}
\end{equation}

According to Eq.~\ref{eq: sup_q_2}, $q^*(\bm{\pi})$ can be computed as follows:

\begin{equation}
    \begin{aligned}
    \log q^*(\bm{\pi}) &= \mathbb{E}_{\bm{Z},\bm{\mu}}[\log p(\bm{Z}, \bm{\pi}, \bm{\mu}, \mathbf{U})] + const \\
    &= \mathbb{E}_{\bm{Z}}[\log p(\bm{Z}|\bm{\pi})] + \log p(\bm{\pi}) + const \\
    &= \sum_{n=1}^N\sum_{k=1}^K \mathbb{E}_{\bm{Z}}[z_{nk}]\log \pi_k + \sum_{k=1}^K (\alpha_o-1)\log \pi_k + const\\
    &= \sum_{k=1}^K\sum_{n=1}^N o_{nk}\log\pi_k + \sum_{k=1}^K (\alpha_o-1)\log \pi_k + const \\
    &= \sum_{k=1}^K (N_k+\alpha_o-1)\log \pi_k +const, \\
    \Longrightarrow q^*(\bm{\pi}) &= \prod_{k=1}^K \pi_k^{N_k+\alpha_o-1} + const \\ 
    &= \prod_{k=1}^K \pi_k^{\alpha_k-1} + const \\
    &= Dir(\bm{\pi}|\bm{\alpha}).
    \end{aligned}
\label{eq: sup_m_step_1}
\end{equation}

Similarly for $q^*(\bm{\mu})$  we can compute it as shown below:
\begin{equation}
    \begin{aligned}
    \log q^*(\bm{\mu}) &= \mathbb{E}_{\bm{Z},\bm{\pi}}[\log p(\bm{Z}, \bm{\pi}, \bm{\mu}, \mathbf{U})] + const \\
    &= \mathbb{E}_{\bm{Z}}[\log p(\mathbf{U}|\bm{Z}, \bm{\mu})] + \log p(\bm{\mu}) + const \\
    &= \sum_{n=1}^N\sum_{k=1}^K \mathbb{E}_{\bm{Z}}[z_{nk}]\log\mathcal{N}(\bm{u}_n|\bm{\mu}_k, \bm{\Lambda}^{-1}) + \sum_{k=1}^K\log \mathcal{N}(\bm{\mu}_k|\bm{m}_o, (\beta_o\bm{\Lambda}^{-1}) + const \\
    &= \frac{1}{2} \sum_{n=1}^N\sum_{k=1}^K o_{nk} \log |\bm{\Lambda}| - \frac{1}{2} \sum_{n=1}^N\sum_{k=1}^K o_{nk} (\bm{u}_n-\bm{\mu}_k)^T\bm{\Lambda}(\bm{u}_n-\bm{\mu}_k) \\
    &+ \frac{1}{2} \sum_{k=1}^K\log |\beta_o\bm{\Lambda}| - \frac{1}{2} \sum_{k=1}^K(\bm{\mu}_k-\bm{m}_o)^T\beta_o\bm{\Lambda}(\bm{\mu}_k-\bm{m}_o).
    \end{aligned}
\label{eq: sup_m_step_2}
\end{equation}

To compute $\beta_k$, we gather the quadratic terms that contain $\bm{\mu}_k$ in Eq.~\ref{eq: sup_m_step_2}:

\begin{equation}
    \begin{aligned}
    (quad) &= -\frac{1}{2}\sum_{n=1}^N\sum_{k=1}^K o_{nk}\bm{\mu}_k^{T}\bm{\Lambda}\bm{\mu}_k -\frac{1}{2}\sum_{k=1}^K\bm{\mu}_k^{T}\beta_o\bm{\Lambda}\bm{\mu}_k \\
    &= -\frac{1}{2}\sum_{k=1}^K\bm{\mu}_k^{T}(N_k\bm{\Lambda}+\beta_o\bm{\Lambda})\bm{\mu}_k \\
    &= -\frac{1}{2}\sum_{k=1}^K\bm{\mu}_k^{T}(\beta_o+N_k)\bm{\Lambda}_k\bm{\mu}_k, \\
    &\Longrightarrow \beta_k = \beta_o + N_k.
    \end{aligned}
\label{eq: sup_m_step_3}
\end{equation}

As for $\bm{m}_k$, we gather the linear terms that contain $\bm{\mu}_k$ in Eq.~\ref{eq: sup_m_step_2}:

\begin{equation}
    \begin{aligned}
    (linear) &= \frac{1}{2}\sum_{n=1}^N\sum_{k=1}^K o_{nk}\bm{\mu}_k^{T}\bm{\Lambda}\bm{u}_n +  \frac{1}{2}\sum_{k=1}^K\bm{\mu}_k^{T}\beta_o\bm{\Lambda}\bm{m}_o \\
    &= \frac{1}{2}\sum_{k=1}^K\bm{\mu}_k^{T}\bm{\Lambda}(\beta_o\bm{m}_o+\sum_{n=1}^N o_{nk}\bm{u}_n) \\
    &= \frac{1}{2}\sum_{k=1}^K\bm{\mu}_k^{T}\beta_k\bm{\Lambda}\bm{m}_k, \\
    &\Longrightarrow \bm{m}_k = \frac{1}{\beta_k}(\beta_o\bm{m}_o + \sum_{n=1}^N o_{nk}\bm{u}_n).
    \end{aligned}
\label{eq: sup_m_step_4}
\end{equation}

Therefore $q^*(\bm{\mu})$ can be reformulated as:
\begin{equation}
    \begin{aligned}
    q^*(\bm{\mu}) = \prod_{k=1}^K q^*(\bm{\mu}_k) = \prod_{k=1}^K\mathcal{N}(\bm{\mu}_k|\bm{m}_k, (\beta_k\bm{\Lambda})^{-1}).
    \end{aligned}
\label{eq: sup_m_step_5}
\end{equation}

We also provide a more detailed calculation of $q^*(\mathbf{Z})$:

\begin{equation}
    \begin{aligned}
    \log q^*(\bm{Z}) &= \mathbb{E}_{\bm{\pi}, \bm{\mu}}[\log p(\bm{Z}, \bm{\pi}, \bm{\mu}, \mathbf{U})] + const \\
    &= \mathbb{E}_{\bm{\pi}}[\log p(\bm{Z}|\bm{\pi})] + \mathbb{E}_{\bm{\mu}}[\log p(\mathbf{U}|\bm{Z}, \bm{\mu})] + const \\
    &= \sum_{n=1}^N\sum_{k=1}^K z_{nk}\left(\mathbb{E}_{\bm{\pi}}[\log\pi_k] + \mathbb{E}_{\bm{\mu}}[\log \mathcal{N}(\bm{u}_n|\bm{\mu}_k, \bm{\Lambda}^{-1})]\right) + const \\
    &= \sum_{n=1}^N\sum_{k=1}^K z_{nk}\log\rho_{nk} + const, 
    \end{aligned}
\label{eq: sup_e_step_1}
\end{equation}

where
\begin{equation}
    \begin{aligned}
    \log\rho_{nk} &= \mathbb{E}_{\bm{\pi}}[\log\pi_k] + \mathbb{E}_{\bm{\mu}}[\log \mathcal{N}(\bm{u}_n|\bm{\mu}_k, \bm{\Lambda}^{-1})] \\
    &= \mathbb{E}_{\bm{\pi}}[\log\pi_k] + \frac{1}{2}\log|\bm{\Lambda}| - \frac{d}{2}\log 2\pi - \frac{1}{2}\mathbb{E}_{\bm{\mu}}[(\bm{u}_n-\bm{\mu}_k)^T\bm{\Lambda}(\bm{u}_n-\bm{\mu}_k)].\\
    \end{aligned}
\label{eq: sup_e_step_2}
\end{equation}

Therefore $q^*(\mathbf{Z})$ can be expressed as: 
\begin{equation}
    \begin{aligned}
    q^*(\bm{Z}) &= \prod_{n=1}^{N} \prod_{k=1}^{K} o_{nk}^{z_{nk}}=\prod_{n=1}^N Categorical(\bm{z}_n|\bm{o}_n), \quad o_{nk} = \frac{\rho_{nk}}{\sum_{j=1}^K \rho_{nj}},
    \end{aligned}
\label{eq: sup_e_step_3}
\end{equation}

we can see that the variable follows a categorical distribution, parameterized by $o_{nk}$, and  $o_{nk}=\mathbb{E}_{\bm{Z}}[z_{nk}]$. As for Eq.~\ref{eq: sup_e_step_2}, more details are shown as follows:

\begin{equation}
    \begin{aligned}
    \mathbb{E}_{\bm{\pi}}[\log\pi_k] &= \psi(\alpha_k)-\psi(\sum_{j=1}^K \alpha_j), \\
    \mathbb{E}_{\bm{\mu}}[(\bm{u}_n-\bm{\mu}_k)^T\bm{\Lambda}(\bm{u}_n-\bm{\mu}_k)] &= \int (\bm{u}_n-\bm{\mu}_k)^T\bm{\Lambda}(\bm{u}_n-\bm{\mu}_k) q^*(\bm{\mu}_k)d\bm{\mu}_k \\
    &= (\bm{u}_n-\bm{m}_k)^T\bm{\Lambda}_k(\bm{u}_n-\bm{m}_k)+\Tr[\bm{\Lambda}\cdot (\beta_k\bm{\Lambda})^{-1}] \\
    &= d\beta_k^{-1}+(\bm{u}_n-\bm{m}_k)^T\bm{\Lambda}(\bm{u}_n-\bm{m}_k),
\end{aligned}
\label{eq: sup_e_step_4}
\end{equation}

$\psi(\cdot)$ is the logarithmic derivative of the gamma function, and the distribution for $\pi_k$ and $\bm{\mu}_k$ follows Eq.~\ref{eq: sup_m_step_1} and~\ref{eq: sup_m_step_5}. Therefore: 
\begin{equation}
    \begin{aligned}
    \log{\rho_{nk}}&=\psi(\alpha_k)-\psi(\sum_{j=1}^K \alpha_j) + \frac{1}{2}\log|\bm{\Lambda}| - \frac{d}{2}\log 2\pi - \frac{1}{2}[d\beta_k^{-1}+(\bm{u}_n-\bm{m}_k)^{T}\bm{\Lambda}(\bm{u}_n-\bm{m}_k)]. \\
    \end{aligned}
\label{eq: sup_e_step_5}
\end{equation}

From the above equations we observe a dependency between priors and posteriors, which can be estimated iteratively depending on the class allocations. Therefore in this paper we propose to solve it under a basic Expectation Maximization framework where we estimate $o_{nk}$ in the E-step, while updating $\alpha_k$, $\beta_k$ and $\bm{m}_k$ in the M-step.

\subsection{Hyperparameter tuning}
\label{tune}

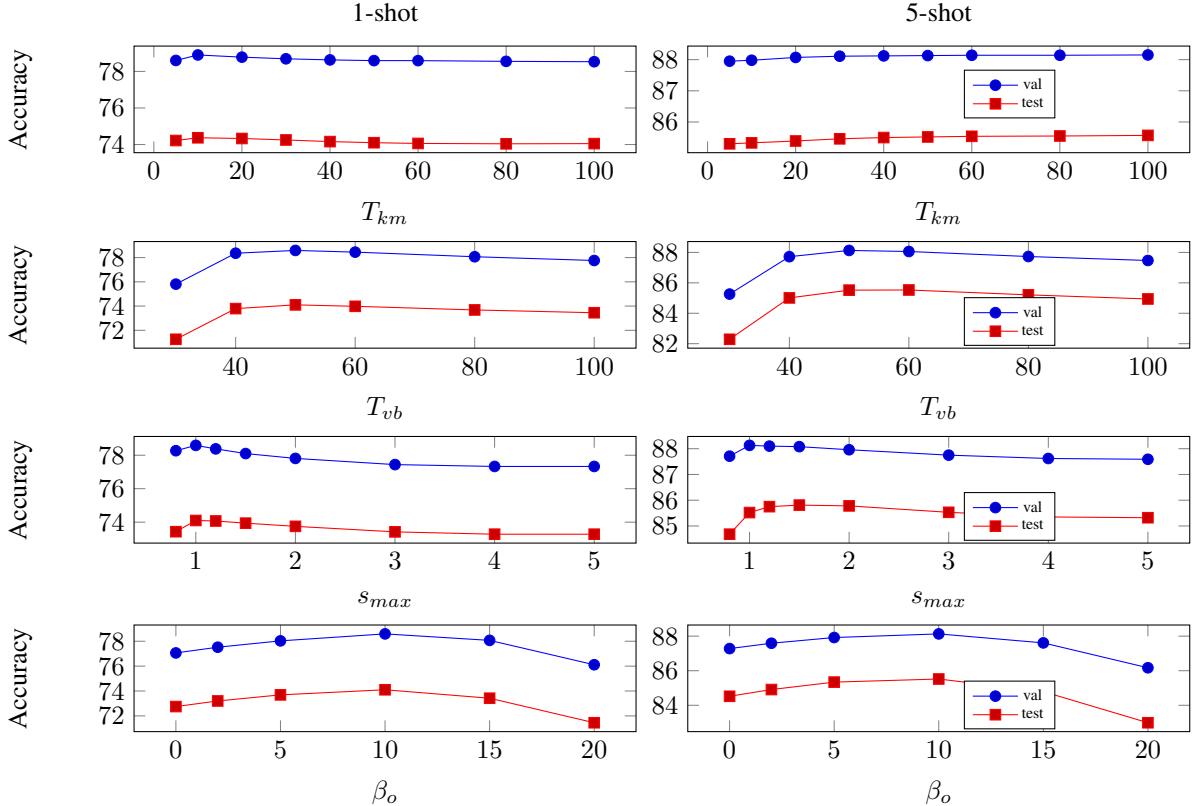
\begin{figure}[h]
    \centering
    \begin{tikzpicture}
       \begin{scope}[]
        \begin{axis}[
            title=1-shot,
            x label style={at={(axis description cs:0.5,0.0)},anchor=north},
            y label style={at={(axis description cs:0.0,.5)},rotate=0,anchor=south},
            xlabel=$T_{km}$,
            ylabel=Accuracy,
            height=3.0cm,
            width=.5\textwidth,
            ]
          
          \addplot coordinates
          {(5, 78.60) (10, 78.90) (20, 78.78) (30, 78.69) (40, 78.63) (50, 78.59) (60, 78.59) (80, 78.55) (100, 78.53)};
          
          \addplot coordinates
          {(5, 74.22) (10, 74.37) (20, 74.33) (30, 74.25) (40, 74.16) (50, 74.10) (60, 74.06) (80, 74.04) (100, 74.05)};
          
        \end{axis}
      \end{scope}
    \end{tikzpicture}
    \begin{tikzpicture}
       \begin{scope}[]
        \begin{axis}[
            title=5-shot,
            x label style={at={(axis description cs:0.5,0.0)},anchor=north},
            y label style={at={(axis description cs:0.0,.5)},rotate=0,anchor=south},
            xlabel=$T_{km}$,
            height=3.0cm,
            width=.5\textwidth,
            legend style={nodes={scale=0.6, transform shape}, at={(0.55, 0.55)}, anchor=west},
            ]
          
          \addlegendentry{val}
          \addplot coordinates
          {(5,87.95) (10,87.98) (20,88.07) (30,88.11) (40,88.12) (50,88.13) (60,88.14) (80,88.14) (100,88.15)};
          
          \addlegendentry{test}
          \addplot coordinates
          {(5,85.30) (10,85.33) (20,85.39) (30,85.46) (40,85.50) (50,85.52) (60,85.54) (80,85.55) (100,85.57)};
          
        \end{axis}
      \end{scope}
    \end{tikzpicture}
    \begin{tikzpicture}
       \begin{scope}[]
        \begin{axis}[
            x label style={at={(axis description cs:0.5,0.0)},anchor=north},
            y label style={at={(axis description cs:0.0,.5)},rotate=0,anchor=south},
            xlabel=$T_{vb}$,
            ylabel=Accuracy,
            height=3.0cm,
            width=.5\textwidth,
            legend style={nodes={scale=0.6, transform shape}, at={(0.55, 0.25)}, anchor=west},
            ]
          
          \addplot coordinates
          {(30, 75.81) (40, 78.36) (50, 78.59) (60, 78.45) (80, 78.07) (100, 77.76)};
          
          \addplot coordinates
          {(30, 71.26) (40, 73.79) (50, 74.10) (60, 73.98) (80, 73.68) (100, 73.45)};
          
        \end{axis}
      \end{scope}
    \end{tikzpicture}
    \begin{tikzpicture}
       \begin{scope}[]
        \begin{axis}[
            x label style={at={(axis description cs:0.5,0.0)},anchor=north},
            y label style={at={(axis description cs:0.0,.5)},rotate=0,anchor=south},
            xlabel=$T_{vb}$,
            height=3.0cm,
            width=.5\textwidth,
            legend style={nodes={scale=0.6, transform shape}, at={(0.55, 0.25)}, anchor=west},
            ]
          
          \addlegendentry{val}
          \addplot coordinates
          {(30,85.26) (40,87.72) (50,88.13) (60,88.06) (80,87.73) (100,87.47)};
          
          \addlegendentry{test}
          \addplot coordinates
          {(30,82.29) (40,85.01) (50,85.52) (60,85.53) (80,85.21) (100,84.94)};
          
        \end{axis}
      \end{scope}
    \end{tikzpicture}
    \begin{tikzpicture}
       \begin{scope}[]
        \begin{axis}[
            x label style={at={(axis description cs:0.5,0.0)},anchor=north},
            y label style={at={(axis description cs:0.0,.5)},rotate=0,anchor=south},
            xlabel=$s_{max}$,
            ylabel=Accuracy,
            height=3.0cm,
            width=.5\textwidth,
            ]
          
          \addplot coordinates
          {(0.8, 78.27) (1, 78.59) (1.2, 78.38) (1.5, 78.10) (2, 77.81) (3,77.44) (4,77.33) (5,77.33)};
          
          \addplot coordinates
          {(0.8, 73.43) (1, 74.10) (1.2, 74.07) (1.5, 73.94) (2, 73.75) (3, 73.42) (4, 73.28) (5,73.28)};
          
        \end{axis}
      \end{scope}
    \end{tikzpicture}
    \begin{tikzpicture}
       \begin{scope}[]
        \begin{axis}[
            x label style={at={(axis description cs:0.5,0.0)},anchor=north},
            y label style={at={(axis description cs:0.0,.5)},rotate=0,anchor=south},
            xlabel=$s_{max}$,
            height=3.0cm,
            width=.5\textwidth,
            legend style={nodes={scale=0.6, transform shape}, at={(0.55, 0.25)}, anchor=west},
            ]
          
          \addlegendentry{val}
          \addplot coordinates
          {(0.8,87.71) (1.0,88.13) (1.2,88.10) (1.5,88.08) (2.0,87.96) (3,87.75) (4,87.62) (5,87.59)};
          
          \addlegendentry{test}
          \addplot coordinates
          {(0.8,84.68) (1.0,85.52) (1.2,85.75) (1.5,85.81) (2.0,85.78) (3, 85.53) (4,85.35) (5,85.32)};
          
        \end{axis}
      \end{scope}
    \end{tikzpicture}
    \begin{tikzpicture}
       \begin{scope}[]
        \begin{axis}[
            x label style={at={(axis description cs:0.5,0.0)},anchor=north},
            y label style={at={(axis description cs:0.0,.5)},rotate=0,anchor=south},
            xlabel=$\beta_o$,
            ylabel=Accuracy,
            height=3.0cm,
            width=.5\textwidth,
            ]
          
          \addplot coordinates
          {(0, 77.06) (2, 77.52) (5, 78.03) (10, 78.59) (15, 78.07) (20, 76.11)};
          
          \addplot coordinates
          {(0, 72.75) (2, 73.20) (5, 73.69) (10, 74.10) (15, 73.42) (20, 71.45)};
          
        \end{axis}
      \end{scope}
    \end{tikzpicture}
    \begin{tikzpicture}
       \begin{scope}[]
        \begin{axis}[
            x label style={at={(axis description cs:0.5,0.0)},anchor=north},
            y label style={at={(axis description cs:0.0,.5)},rotate=0,anchor=south},
            xlabel=$\beta_o$,
            height=3.0cm,
            width=.5\textwidth,
            legend style={nodes={scale=0.6, transform shape}, at={(0.55, 0.25)}, anchor=west},
            ]
          
          \addlegendentry{val}
          \addplot coordinates
          {(0, 87.28) (2, 87.59) (5, 87.92) (10, 88.13) (15, 87.61) (20, 86.17)};
          
          \addlegendentry{test}
          \addplot coordinates
          {(0, 84.52) (2, 84.91) (5, 85.34) (10, 85.52) (15, 84.79) (20, 82.99)};
          
        \end{axis}
      \end{scope}
    \end{tikzpicture}
\caption{Hyperparameter tuning of our proposed method. Here we tune 4 hyperparameters of BAVARDAGE on \textit{mini}-Imagenet (backbone: WRN) in the unbalanced setting.}
  \label{fig:sub_tuning}
\end{figure}

In this section we detail about how the hyperparameters in our proposed method are obtained. Namely, for a standard Few-Shot benchmark that has been split into base-validation-novel class set, we firstly tune our model using validation set and choose the hyperparameters accordingly before applying to the novel set. For example in Figure~\ref{fig:sub_tuning} we tune two temperature parameters $T_{km}$, $T_{vb}$, the scaling up-bound parameter $s_{max}$ and the VB prior $\beta_o$ that are used in our proposed BAVARDAGE. The blue curves show the performance on validation set while the red curves show the accuracy on the novel set (benchmark: \textit{mini}-Imagenet). From the figure we see a similar behavior between two sets in terms of performance, $T_{km}$ has little impact on the accuracy, same for $T_{vb}$ when it is large. For $s_{max}$ we observe an uptick when it is around 1, followed by a slowing decrease and finally stabilizing to the same accuracy when it becomes larger. In this paper we tune hyperparameters for each benchmark in the same way. For \textit{tiered}-Imagenet we set $T_{km}$, $T_{vb}$ and $s_{max}$ to be 10, 100, 2 in the balanced setting, 100, 100, 1 in the unbalanced setting; for CUB we set them to be 10, 4, 5 in both balanced and unbalanced settings; and for FC100 and CIFAR-FS we set the hyperparameters to be the same as \textit{mini}-Imagenet. As for $\beta_o$ we set it to be 10 across datasets since it gives the best performance.

\subsection{Additional experiments on other Few-Shot benchmarks}
\label{results}

\begin{table}[t]
  \caption{Detailed results of BAVARDAGE with confidence interval of $95\%$ on the Few-Shot benchmarks, along with a baseline accuracy using Soft-KMEANS. We use RN18 and WRN pretrained from~\cite{veilleux2021realistic}, RN12 and RN12* pretrained from~\cite{bendou2022easy}. }
  \label{results_2}
  \centering
  \resizebox{0.9\columnwidth}{!}{
  \begin{tabular}{lccccc}
    \toprule
    \textbf{\textit{mini}-Imagenet} & & \multicolumn{2}{c}{\textbf{unbalanced}} & \multicolumn{2}{c}{\textbf{balanced}} \\
    \cmidrule(r){3-6}
    Method & Backbone & \textbf{1-shot} & \textbf{5-shot} & \textbf{1-shot} & \textbf{5-shot} \\
    \midrule
    \multirow{4}{*}{Soft-KMEANS} & RN18~\cite{veilleux2021realistic} & $68.82\pm 0.27\%$ & $81.27\pm 0.17\%$ & $73.47\pm 0.26\%$ & $83.04\pm 0.15\%$\\
     & WRN~\cite{veilleux2021realistic} & $71.35\pm 0.27\%$ & $82.41\pm 0.16\%$ & $75.70\pm 0.25\%$ & $84.42\pm 0.14\%$\\
     & RN12~\cite{bendou2022easy} & $75.65\pm 0.25\%$ & $86.35\pm 0.14\%$ & $80.81\pm 0.24\%$ & $87.92\pm 0.12\%$\\
     & RN12*~\cite{bendou2022easy} & $77.51\pm 0.26\%$ & $87.78\pm 0.14\%$ & $82.14\pm 0.24\%$ & $89.08\pm 0.12\%$\\
    \midrule
    \multirow{4}{*}{BAVARDAGE} & RN18~\cite{veilleux2021realistic} & $71.01\pm 0.31\%$ & $83.60\pm 0.17\%$ & $75.07\pm 0.28\%$ & $84.49\pm 0.14\%$\\
     & WRN~\cite{veilleux2021realistic} & $74.10\pm 0.30\%$ & $85.52\pm 0.16\%$ & $78.51\pm 0.27\%$ & $87.41\pm 0.13\%$\\
     & RN12~\cite{bendou2022easy} & $77.85\pm 0.28\%$ & $88.02\pm 0.14\%$ & $82.67\pm 0.25\%$ & $89.50\pm 0.11\%$\\
     & RN12*~\cite{bendou2022easy} & $79.76\pm 0.29\%$ & $89.85\pm 0.13\%$ & $84.80\pm 0.25\%$ & $91.65\pm 0.10\%$\\
    \bottomrule
    
    \textbf{\textit{tiered}-Imagenet} & & \multicolumn{2}{c}{\textbf{unbalanced}} & \multicolumn{2}{c}{\textbf{balanced}} \\
    \cmidrule(r){3-6}
    Method & Backbone & \textbf{1-shot} & \textbf{5-shot} & \textbf{1-shot} & \textbf{5-shot} \\
    \midrule
    \multirow{4}{*}{Soft-KMEANS} & WRN~\cite{veilleux2021realistic} & $73.92\pm 0.28\%$ & $85.02\pm 0.18\%$ & $78.59\pm 0.27\%$ & $85.76\pm 0.16\%$\\
     & RN18~\cite{veilleux2021realistic} & $73.79\pm 0.28\%$ & $84.65\pm 0.18\%$ & $78.34\pm 0.27\%$ & $85.52\pm 0.17\%$\\
     & RN12~\cite{bendou2022easy} & $78.15\pm 0.27\%$ &  $87.65\pm 0.17\%$ & $83.11\pm 0.25\%$ & $88.80\pm 0.15\%$\\
     & RN12*~\cite{bendou2022easy} & $79.62\pm 0.27\%$& $88.61\pm 0.16\%$& $84.08\pm 0.24\%$& $89.56\pm 0.14\%$\\
    \midrule
    \multirow{4}{*}{BAVARDAGE} & WRN~\cite{veilleux2021realistic} & $77.45\pm 0.31\%$ & $87.48\pm 0.18\%$ & $81.47\pm 0.28\%$ & $88.27\pm 0.16\%$\\
     & RN18~\cite{veilleux2021realistic} & $76.55\pm 0.31\%$ & $86.46\pm 0.19\%$ & $80.32\pm 0.28\%$ & $87.14\pm 0.16\%$\\
     & RN12~\cite{bendou2022easy} & $79.38\pm 0.29\%$ &  $88.04\pm 0.18\%$ & $83.52\pm 0.26\%$ & $89.03\pm 0.15\%$\\
     & RN12*~\cite{bendou2022easy} & $81.17\pm 0.29\%$& $89.63\pm 0.17\%$& $85.20\pm 0.25\%$& $90.41\pm 0.14\%$ \\
    \bottomrule
    
    \textbf{CUB} & & \multicolumn{2}{c}{\textbf{unbalanced}} & \multicolumn{2}{c}{\textbf{balanced}} \\
    \cmidrule(r){3-6}
    Method & Backbone & \textbf{1-shot} & \textbf{5-shot} & \textbf{1-shot} & \textbf{5-shot} \\
    \midrule
    \multirow{3}{*}{Soft-KMEANS} & RN18~\cite{veilleux2021realistic} & $77.54\pm 0.26\%$ & $86.70\pm 0.14\%$ & $82.67\pm 0.24\%$ & $89.04\pm0.11\%$\\
     & RN12~\cite{bendou2022easy} & $81.24\pm 0.25\%$ &  $87.27\pm 0.14\%$ & $84.87\pm 0.22\%$ & $89.64\pm 0.11\%$\\
     & RN12*~\cite{bendou2022easy} & $82.40\pm 0.24\%$ & $89.40\pm 0.13\%$ &$87.38\pm 0.20\%$ & $91.29\pm 0.10\%$\\
    \midrule
    \multirow{3}{*}{BAVARDAGE} & RN18~\cite{veilleux2021realistic} & $82.00\pm 0.28\%$ & $90.67\pm 0.12\%$ & $85.64\pm 0.25\%$ & $91.42\pm 0.10\%$\\
     & RN12~\cite{bendou2022easy} & $83.12\pm 0.26\%$ &  $90.81\pm 0.12\%$ & $87.41\pm 0.22\%$ & $92.03\pm 0.09\%$\\
     & RN12*~\cite{bendou2022easy} & $86.96\pm 0.24\%$ & $92.84\pm 0.10\%$& $90.42\pm 0.20\%$ & $93.50\pm 0.08\%$\\
    \bottomrule
    
    \textbf{FC100} & & \multicolumn{2}{c}{\textbf{unbalanced}} & \multicolumn{2}{c}{\textbf{balanced}} \\
    \cmidrule(r){3-6}
    Method & Backbone & \textbf{1-shot} & \textbf{5-shot} & \textbf{1-shot} & \textbf{5-shot} \\
    \midrule
    \multirow{2}{*}{Soft-KMEANS} & RN12~\cite{bendou2022easy} & $51.24\pm 0.27\%$ & $64.70\pm 0.22\%$ & $54.59\pm 0.26\%$ & $66.37\pm 0.20\%$\\
     & RN12*~\cite{bendou2022easy} & $51.64\pm 0.27\%$ & $65.26\pm 0.22\%$ & $54.87\pm 0.26\%$ & $66.89\pm 0.20\%$ \\
    \midrule
    \multirow{2}{*}{BAVARDAGE} & RN12~\cite{bendou2022easy} & $52.60\pm 0.32\%$ & $65.35\pm 0.25\%$ & $56.66\pm 0.28\%$ & $69.69\pm 0.21\%$\\
     & RN12*~\cite{bendou2022easy} & $53.78\pm 0.30\%$ & $68.75\pm 0.24\%$ & $57.27\pm 0.29\%$ & $70.60\pm 0.21\%$ \\
    \bottomrule
    \textbf{CIFAR-FS} & & \multicolumn{2}{c}{\textbf{unbalanced}} & \multicolumn{2}{c}{\textbf{balanced}} \\
    \cmidrule(r){3-6}
    Method & Backbone & \textbf{1-shot} & \textbf{5-shot} & \textbf{1-shot} & \textbf{5-shot} \\
    \midrule
    \multirow{2}{*}{Soft-KMEANS} & RN12~\cite{bendou2022easy} & $80.72\pm 0.25\%$ & $88.31\pm 0.17\%$ & $85.47\pm 0.22\%$ & $89.36\pm 0.15\%$\\
     & RN12*~\cite{bendou2022easy} & $81.75\pm 0.25\%$ & $88.92\pm 0.17\%$ & $86.07\pm 0.22\%$ & $89.85\pm 0.15\%$ \\
    \midrule
    \multirow{2}{*}{BAVARDAGE} & RN12~\cite{bendou2022easy} & $82.68\pm 0.27\%$ & $88.97\pm 0.18\%$ & $86.20\pm 0.23\%$ & $89.58\pm 0.15\%$\\
     & RN12*~\cite{bendou2022easy} & $83.82\pm 0.27\%$ & $89.84\pm 0.18\%$ & $87.35\pm 0.23\%$ & $90.63\pm 0.16\%$\\
    \bottomrule
  \end{tabular}
  }
\end{table}

In Section 4 in the paper we tested our proposed method on three standard Few-Shot benchmarks: \textit{mini}-Imagenet\footnote{\url{https://github.com/yaoyao-liu/mini-imagenet-tools}}, \textit{tiered}-Imagenet\footnote{\url{https://github.com/yaoyao-liu/tiered-imagenet-tools}} and CUB\footnote{\url{http://www.vision.caltech.edu/datasets/cub\_200\_2011}}, following the same setting as presented in~\url{https://github.com/oveilleux/Realistic_Transductive_Few_Shot}. In this section we further conduct experiments on two other well-known Few-Shot datasets: 1) FC100 (\url{https://github.com/ElementAI/TADAM}) is a recent split dataset based on CIFAR-100~\cite{krizhevsky2009learning} that contains 60 base classes for training, 20 classes for validation and 20 novel classes for evaluation, each class is composed of 600 images of size 32x32 pixels; 2) CIFAR-FS (\url{https://github.com/bertinetto/r2d2}) is also sampled from CIFAR-100 and shares the same quantity of classes in the base-validation-novel splits as for \textit{mini}-Imagenet. Each class contains 600 images of size 32x32 pixels. In Table~\ref{results_2} below we report the accuracy of our proposed method on all benchmarks, note that for FC100 and CIFAR-FS we believe to be among the first to conduct experiments in the unbalanced setting. 

In Table~\ref{results_2} we also show the results using WRN and RN18 pretrained from~\cite{veilleux2021realistic} and RN12 pretrained from~\cite{bendou2022easy}, same as Table 1 in the paper, with a confidence interval of $95\%$ added next to the accuracy. In addition, given that some works~\cite{luo2021rectifying, zhang2020deepemd} in the field utilize data augmentation techniques to extract features based on images in  original dimensions instead of reduced ones, here we apply our BAVARDAGE following the same setting and report the accuracy on a pretrained RN12 feature extractor~\cite{bendou2022easy} with data augmentation (denote RN12*). For comparison purpose we also provide a baseline accuracy on each Few-Shot benchmark using Soft-KMEANS algorithm. 

With BAVARDAGE, we observe a clear increase of accuracy for all datasets compared with Soft-KMEANS in both balanced and unbalanced settings, suggesting the genericity of the proposed method. As for the computational time, we evaluate an average of $1.72$ seconds per accuracy (on 10,000 Few-Shot tasks) using a GeForce RTX 3090 GPU.

\clearpage
{
\small
\bibliographystyle{abbrv}

}


\end{document}